\documentclass[runningheads]{llncs}
\usepackage[T1]{fontenc}

\usepackage{graphicx}%
\usepackage{multirow}%
\usepackage{amsmath,amssymb,amsfonts}%
\usepackage{hyperref}
\usepackage{mathrsfs}%
\usepackage[title]{appendix}%
\usepackage{xcolor}%
\usepackage{textcomp}%
\usepackage{manyfoot}%
\usepackage{booktabs}
\usepackage{algorithm}%
\usepackage{algorithmicx}%
\usepackage{algpseudocode}%
\usepackage{listings}%
\usepackage[numbers]{natbib}
\usepackage{paralist}

\usepackage{multirow}
\usepackage[normalem]{ulem}
\useunder{\uline}{\ul}{}
\usepackage[utf8]{inputenc}

\usepackage{wrapfig}

\begin{document}

\title{TrOCR for Medieval HTR: A Systematic Ablation Study with Cross-Dataset Validation}

\author{Sachin Sharma\orcidID{0000-0002-9918-7914} \and Michele Flammini\orcidID{0000-0003-0327-3728} \and Federico Simonetta\orcidID{0000-0002-5928-9836}}

\authorrunning{S. Sharma \& F. Simonetta}

\institute{GSSI -- Gran Sasso Science Institute\\ L'Aquila, Italy\\
	\email{\{name\}.\{surname\}@gssi.it}}


\maketitle

\begin{abstract}
Fine-tuning transformer-based handwritten text recognition (HTR) models on medieval manuscripts is challenging because these models are pre-trained on modern text and must adapt to a very different visual domain. This paper studies how three controllable fine-tuning choices (contrast normalization, data augmentation, and layer freezing) affect recognition accuracy when adapting TrOCR to small historical datasets.
We run controlled experiments on a 13th-century Italian manuscript (I-CT 91 ``Cortonese'') and replicate the same experimental grid on the public READ-16 benchmark as robustness evidence. On Cortonese, our best configuration achieves 8.03\% character error rate (CER). Statistical comparisons across 13 configurations show that freezing up to three encoder layers or six decoder layers does not significantly harm accuracy, while deeper freezing becomes progressively detrimental. Removing contrast normalization (CLAHE) yields 7.84\% CER, comparable to a domain-specialized baseline, suggesting strong optimization can reduce reliance on image preprocessing. Cross-dataset validation on READ-16 shows that decoder freezing thresholds transfer more robustly than encoder thresholds, and combined freezing strategies require dataset-specific re-validation. Finally, we use Grad-CAM gradient attributions and decoder cross-attention maps to diagnose error patterns and failure modes revealed by the ablations.
	Source code is available at \url{https://github.com/LaudareProject/TrOCR-analysis}
	

\end{abstract}

\section{Introduction}

Handwritten text recognition (HTR) is a prerequisite for turning medieval manuscripts into searchable, analyzable corpora, yet performance remains fragile. Relative to modern printed documents, manuscripts combine shifting letterforms, pervasive abbreviations, degraded supports (faded ink, bleed-through, staining), and irregular layouts~\cite{simonetta2024optical}. Systems tuned to clean print therefore generalize poorly in this domain~\cite{semanticscholar32,semanticscholar41}.

Recent transformer-based HTR models have improved modern handwriting recognition. TrOCR~\cite{li2021trocr}, for instance, performs strongly in contemporary domains, but medieval material lies far from its pre-training distribution in handwriting style, orthographic conventions, and linguistic structure~\cite{semanticscholar14,semanticscholar31}. We select TrOCR as the baseline because it is a strong and widely used encoder--decoder HTR model with public checkpoints, making its fine-tuning behavior practically relevant. At the same time, historical collections often provide only limited supervision (typically a few thousand labeled lines), so the details of fine-tuning can dominate the final outcome~\cite{semanticscholar20,semanticscholar46}.

The historical HTR literature includes a wide range of application-focused studies~\cite{Toselli2018READ,Fischer2011SaintGall,Toselli2015Bentham}, and recent work confirms that transformer-based models can be adapted to historical documents~\cite{strobel2022transformerbasedhtrhistoricaldocuments,Xu_2020,Fischer2023,Kim2021Donut}. However, reported gains are often hard to interpret because preprocessing, augmentation, and optimization are frequently adjusted together; controlled ablations remain relatively uncommon~\cite{semanticscholar46}. It is also unclear how far a single hyperparameter configuration transfers across scripts, languages, and periods.

Interpretability is similarly underexplored. Attention visualizations are a common starting point; Donut, for example, reports decoder cross-attention maps with emergent localization behavior~\cite{semanticscholar110}. Yet attention weights are not explanations by default~\cite{chefer2021transformer}. Gradient-based attributions such as Grad-CAM~\cite{selvarajuGradcamVisualExplanations2017} are more tightly coupled to the training objective, but attention- and gradient-derived signals are still rarely used as quantitative diagnostics in OCR/HTR~\cite{semanticscholar110,semanticscholar111}.

Within Document Analysis Systems, this work addresses the line-level HTR component of a broader transcription pipeline: manuscript image $\rightarrow$ segmentation $\rightarrow$ line-level HTR $\rightarrow$ correction/review $\rightarrow$ document-level transcription use.

We study this question with controlled ablations that isolate preprocessing, data augmentation, and encoder/decoder layer freezing. Overall, this work contributes practical recommendations for TrOCR users:

\begin{enumerate}
	\item We provide controlled ablations on a 13th-century Italian manuscript, isolating preprocessing, augmentation, and encoder--decoder freezing effects on CER and generalization, with accompanying statistical significance tests.
	\item We provide a full 13-configuration ablation replicated on READ-16 with line-level statistical validation, quantifying which findings transfer across scripts, languages, and centuries and which require dataset-specific re-validation.\footnote{The source code used for the experiments is available at \url{https://github.com/LaudareProject/TrOCR-analysis}. The dataset is currently under review and will be released upon completion of the review.}
    \item a qualitative diagnostic analysis using Grad-CAM and cross-attention to interpret ablation-linked error patterns and failure modes.
\end{enumerate}

In order to test whether fine-tuning conclusions are specific to a single corpus or transfer across typologically distant collections, we perform a cross-dataset validation using two datasets that differ along several complementary axes: script, language, time period, and document type (I-CT 91 and READ-16).

We do not claim to outperform all existing HTR systems. Instead, we provide a  controlled ablation study and actionable fine-tuning guidance for practitioners working with small medieval HTR datasets.

\section{Dataset and Experimental Setup}

\subsection{Dataset}
\begin{figure}
	\centering
	\includegraphics[width=0.5\linewidth]{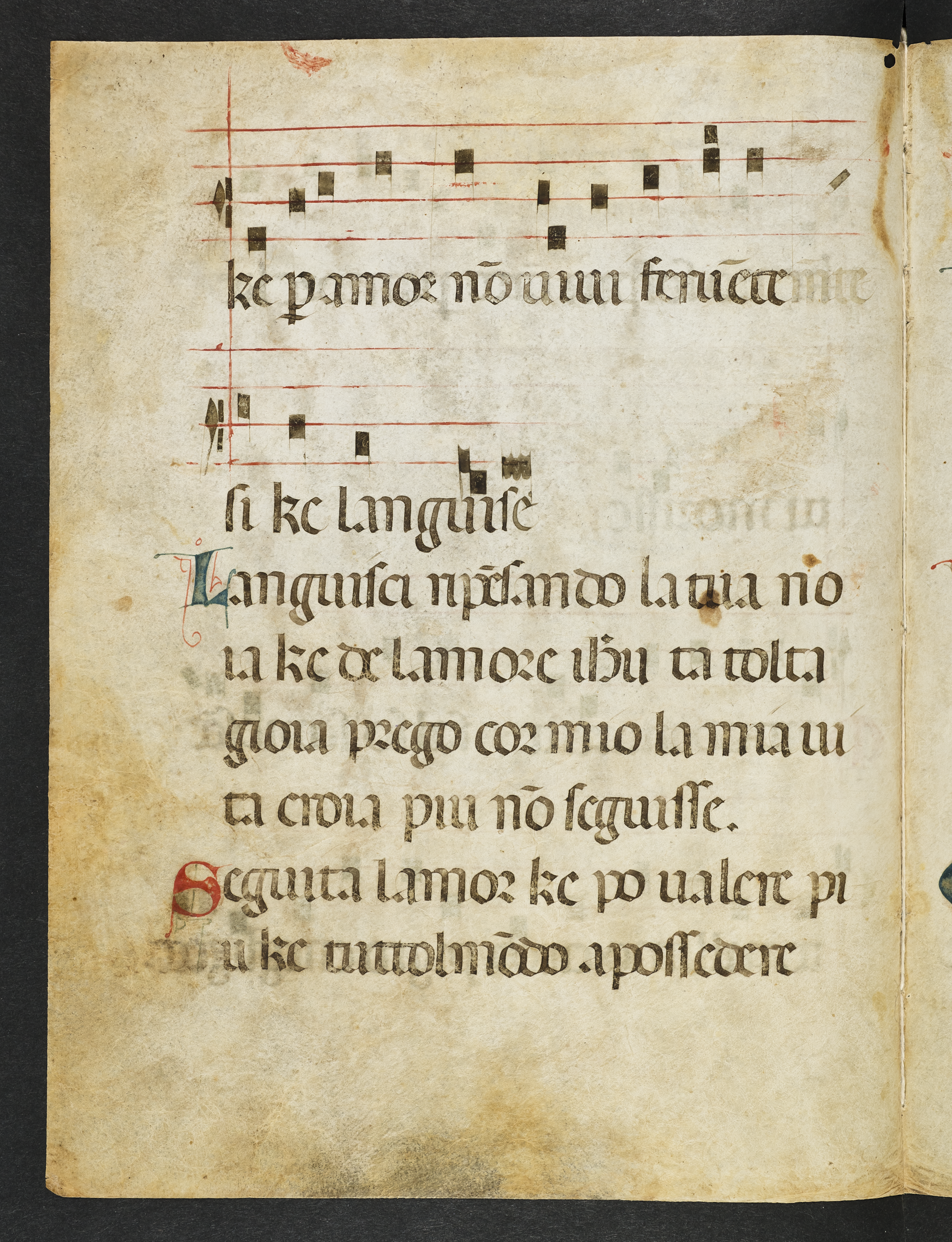}
	\caption{Example of a page from the I-CT 91 ``Cortonese'' manuscript}
	\label{fig:manuscript}
\end{figure}

All experiments were performed on the I-CT 91 ``Cortonese'' manuscript (Fig.~\ref{fig:manuscript})~\cite{sharma2025experimenting}, an Italian laude source dating from the 13th century. The dataset comprises 259 manuscript pages, 2,517 annotated text lines, 14,214 words, and 75,667 characters. Each line is provided with verified ground truth transcriptions. The average image resolution is 2082$\times$2717 with 300 DPI and average text line size of 1352.85$\times$159.64. The manuscript presents multiple paleographic challenges, including the presence of two different scribal hands, the coexistence of textual and neumatic notation, and substantial physical degradation such as faded ink and parchment staining. These characteristics make the collection low-resource medieval test case for line-level HTR adaptation under real-world manuscript conditions.

To ensure reproducibility and fair evaluation, we adopted one of the standardized splits provided by our project. The partition preserves page-level separation to prevent data leakage between training and testing subsets. The dataset includes 1{,}666 text lines for training ($66.2\%$), 397 for validation ($15.8\%$), and 454 for testing ($18\%$). Unlike random splitting, this configuration guarantees consistent comparisons across studies and future reproducibility for subsequent research on the same material. The exact split indices and all experimental source code are publicly available in the code repository to facilitate replication once the dataset is released.

\section{Methodology}


In order to define a rigorous protocol, we define the main set the following research question: \textit{when adapting a pre-trained TrOCR model to a small medieval HTR dataset, which architectural choices impact the CER metric the most, and to which extent the impact is measurable?} 

To investigate this question, we examine three controllable fine-tuning levers: \begin{inparaenum}[\itshape a)\upshape]
    \item preprocessing,
    \item data augmentation, and
    \item parameter-efficient fine-tuning through layer freezing.
\end{inparaenum}
These axes were selected because they are common, controllable design decisions in low-resource HTR pipelines and directly trade off accuracy, compute, and manual pipeline effort.

All other training factors are held constant throughout the experiments. The experimental design modifies one lever at a time relative to a fixed full fine-tuning baseline, thereby enabling a clean attribution of performance differences to individual methodological choices.

\paragraph{Preprocessing Pipeline:}
The preprocessing pipeline was designed to enhance image readability and improve OCR robustness under varying manuscript conditions. Contrast enhancement was performed using Contrast Limited Adaptive Histogram Equalization (CLAHE) with a clip limit of 2.0 and an 8$\times$8 tile grid~\cite{10.5555/180895.180940,Pizer1987AdaptiveHE}. For comparison, standard histogram equalization was also considered during first experimentation. Deskewing was applied using a projection profile method constrained within $\pm5^\circ$, ensuring consistent text line orientation. Finally, images were normalized to a fixed height of 64 pixels and standardized according to ImageNet statistics.

To improve robustness, several augmentation techniques were applied during training. Geometric transformations included random rotations within $\pm5^\circ$ (probability 0.4) and elastic deformations ($\alpha = 20$, $\sigma = 5$, $p = 0.25$). Photometric augmentations involved Gaussian blur ($p = 0.35$), brightness adjustment ($p = 0.40$), and contrast variation ($p = 0.30$). Additionally, manuscript degradation effects were simulated through speckle noise ($p = 0.25$), morphological operations ($p = 0.15$), and random shadow or staining overlays ($p = 0.15$). These augmentations emulate common challenges in medieval texts such as uneven illumination, ink bleed, and parchment defects.

\paragraph{Ablation Experiments:}
We run controlled ablations over three levers: preprocessing (CLAHE), data augmentation, and parameter-efficient fine-tuning via layer freezing. We keep the dataset split, evaluation protocol, and TrOCR-base initialization fixed, and change one lever at a time. For preprocessing and augmentation, we compare four variants under full fine-tuning (\texttt{enc\_0\_dec\_0}): (i) CLAHE + augmentation (baseline), (ii) no CLAHE, (iii) no augmentation, and (iv) no CLAHE + no augmentation. For layer freezing, we sweep $K \in \{0,3,6,9,12\}$ frozen layers in the encoder or decoder while keeping the other module fully trainable. Because both modules have 12 layers, this corresponds to 25\% steps (0\%--100\% frozen). We then test \texttt{enc\_3\_dec\_6} to verify whether the individually non-significant freezing thresholds remain non-significant when applied jointly.

Following \citeauthor{huttner2025lowrank}, we train with AdamW and a One-Cycle schedule for 50 epochs (no early stopping). The learning rate starts at $1\times 10^{-9}$, rises to $5.5\times 10^{-6}$ over the first five epochs (10\% of the run), and then anneals to a value lower than the peak by a factor of $2.2\times 10^{4}$. Momentum follows the same One-Cycle schedule (0.85--0.95). We regularize with weight decay (0.01) and label smoothing (0.1). We train with 8 samples per device and accumulate gradients for two steps, for an effective batch size of 16.

\paragraph{Kraken Baseline:}
In order to obtain a domain-specific baseline, we fine-tuned a CNN model using Kraken~\cite{kiessling2026version}. We used the pre-trained CATMuS Medieval model (version 1.6.0)~\cite{pinche_2024_12743230}. This model was trained using the default Kraken architecture on a large dataset comprising 160 000 text lines and 5 millions characters from documents from the 8th to the 16th centuries~\cite{clerice2024catmus}. Importantly, this dataset included various European languages, including Italian.
We fine-tuned the CATMuS model on our I-CT 91 using early stopping with patience of 15 epochs and the augmentation strategy implemented by Kraken.\footnote{For more info about the augmentation pipeline, see the Python module \texttt{kraken/lib/dataset/recognition.py} as implemented in version 5.3.0.}

\paragraph{Evaluation Metrics:}
Model performance was evaluated using the standard Character Error Rate (CER) and Word Error Rate (WER) metrics. The Character Error Rate is defined as:
\[
	CER = \frac{S + D + I}{N},
\]
where $S$, $D$, and $I$ represent the number of substitutions, deletions, and insertions respectively, and $N$ is the length of the reference transcription. The Word Error Rate follows the same formulation but is computed at the word level. Both metrics were used to provide a comprehensive assessment of transcription accuracy at fine-grained and semantic levels.

\section{Results}

\begin{table}[]
	\caption{I-CT 91 Ablation Study}
	\label{tab:ict91_ablation}
	\begin{center}
		\begin{tabular}{lcccccc}
			\toprule
			\textbf{Configuration}         & \textbf{\begin{tabular}[c]{@{}c@{}}Frozen \\ (enc/dec)\end{tabular}} & \textbf{\begin{tabular}[c]{@{}c@{}}Trainable \\ Params\end{tabular}} & \textbf{CER} & \textbf{WER} & \textbf{\begin{tabular}[c]{@{}c@{}}mean \\ $\Delta$CER\end{tabular}} & \textbf{Sign.} \\ 
            \midrule
			enc\_0\_dec\_0 $\star$ & 0 / 0                                                                & 333.9M (100\%)                                                       & 8.0          & 31.9         & ---                                                                  & ---            \\
			No Augm                        & 0 / 0                                                                & 333.9M (100\%)                                                       & 8.4          & 32.1         & +0.45                                                                & ns             \\
			No CLAHE No Augm               & 0 / 0                                                                & 333.9M (100\%)                                                       & 8.2          & 32.3         & +0.18                                                                & ns             \\
			No CLAHE                       & 0 / 0                                                                & 333.9M (100\%)                                                       & \textbf{7.8}          & \textbf{31.1}         & -0.19                                                                & ns             \\ 
            
			enc\_3\_dec\_0                 & 3 / 0                                                                & $\sim$312.7M (93.6\%)                                                & 8.2          & 32.4         & +0.20                                                                & ns             \\
			enc\_6\_dec\_0                 & 6 / 0                                                                & $\sim$291.4M (87.3\%)                                                & 9.7          & 34.8         & +1.66                                                                & ***            \\
			enc\_9\_dec\_0                 & 9 / 0                                                                & $\sim$270.2M (80.9\%)                                                & 13.8         & 42.5         & +5.79                                                                & ***            \\
			enc\_12\_dec\_0                & 12 / 0                                                               & 248.9M (74.5\%)                                                      & 16.5         & 47.5         & +8.43                                                                & ***            \\
            
			enc\_0\_dec\_3                 & 0 / 3                                                                & $\sim$285.1M (85.4\%)                                                & 8.2          & 32.0         & +0.18                                                                & ns             \\
			enc\_0\_dec\_6                 & 0 / 6                                                                & $\sim$236.3M (70.8\%)                                                & 8.4          & 32.0         & +0.34                                                                & ns             \\
			enc\_0\_dec\_9                 & 0 / 9                                                                & $\sim$187.5M (56.1\%)                                                & 8.9          & 33.4         & +0.84                                                                & *              \\
			enc\_0\_dec\_12                & 0 / 12                                                               & 138.7M (41.5\%)                                                      & 10.3         & 37.4         & +2.23                                                                & ***            \\
            
			enc\_3\_dec\_6                 & 3 / 6                                                                & $\sim$215M (64.4\%)                                                  & 8.6          & 32.6         & +0.52                                                                & ns             \\
            
		      \midrule                                                                                            
			Kraken-CATMuS (Baseline)       & ---                                                                  & ---                                                                  & 7.9          & 31.2         & -0.1                                                                 & ---             \\
            \bottomrule
		\end{tabular}
	\end{center}
	\vspace{2mm}
	\footnotesize{
		\textbf{Notes:}
		Sign. = Wilcoxon signed-rank test vs baseline, Holm-corrected, computed between line-wise CER ($N=454$ lines). Friedman omnibus test: $\chi^2=1001.19$, $p < 10^{-151}$. $\star$ marks the full fine-tuning.
		*** $p < 0.001$, ** $p < 0.01$, * $p < 0.05$, ns = not significant ($p_{\text{holm}} \ge 0.05$). Kraken–CATMuS is an external domain-specialized baseline included for context; it is not a TrOCR variant and is not part of the statistical comparisons.
	}
\end{table}

\begin{table*}[t]
	{
    \centering
	\caption{READ-16 Ablation Study}
	\label{tab:read16_ablation}
	\begin{tabular}{lcccccc}
		\toprule
		\textbf{Configuration}         & \textbf{\begin{tabular}[c]{@{}c@{}}Frozen \\ (enc/dec)\end{tabular}} & \textbf{\begin{tabular}[c]{@{}c@{}}Trainable \\ Params\end{tabular}} & \textbf{CER} & \textbf{WER} & \textbf{\begin{tabular}[c]{@{}c@{}}mean \\ $\Delta$CER\end{tabular}} & \textbf{Sign.} \\ 
		\midrule
		enc\_0\_dec\_0 $\star$                           & 0 / 0            & 333.9M (100\%)        & \textbf{5.16}  & \textbf{21.18} & --          & --    \\

		No Augm                                          & 0 / 0            & 333.9M (100\%)        & 5.92  & 23.97 & +0.75       & **    \\
		No CLAHE No Augm                                 & 0 / 0            & 333.9M (100\%)        & 5.62  & 23.24 & +0.45       & ns    \\
		No CLAHE                                         & 0 / 0            & 333.9M (100\%)        & 5.30  & 21.91 & +0.14       & ns    \\

		enc\_3\_dec\_0                                   & 3 / 0            & $\sim$312.7M (93.6\%) & 5.94  & 23.86 & +0.78       & *     \\
		enc\_6\_dec\_0                                   & 6 / 0            & $\sim$291.4M (87.3\%) & 7.37  & 25.86 & +2.20       & ***   \\
		enc\_9\_dec\_0                                   & 9 / 0            & $\sim$270.2M (80.9\%) & 11.80 & 33.25 & +6.63       & ***   \\
		enc\_12\_dec\_0                                  & 12 / 0           & 248.9M (74.5\%)       & 15.71 & 39.82 & +10.54      & ***   \\

		enc\_0\_dec\_3                                   & 0 / 3            & $\sim$285.1M (85.4\%) & 5.60  & 22.45 & +0.44       & ns    \\
		enc\_0\_dec\_6                                   & 0 / 6            & $\sim$236.3M (70.8\%) & 5.69  & 22.14 & +0.53       & ns    \\
		enc\_0\_dec\_9                                   & 0 / 9            & $\sim$187.5M (56.1\%) & 5.96  & 24.00 & +0.80       & **    \\
		enc\_0\_dec\_12                                  & 0 / 12           & 138.7M (41.5\%)       & 8.14  & 31.78 & +2.97       & ***   \\

		enc\_3\_dec\_6                                   & 3 / 6            & $\sim$215M (64.4\%)   & 6.49  & 24.37 & +1.33       & ***   \\

		\midrule
		VAN \cite{coquenet2023endtoend}                  & --               & --                    & 4.10  & 16.29 & --          & --    \\
		GatedLexiconNet \cite{kumari2024gatedlexiconnet} & --               & --                    & 2.13  & 6.52  & --          & --    \\
		TrOCR base \cite{huttner2025lowrank}             & --               & --                    & 4.81  & 23.22 & --          & --    \\
		\bottomrule
	\end{tabular}

	\vspace{0.5em}
    }
	\footnotesize{
		Sign. = Wilcoxon signed-rank test vs enc\_0\_dec\_0 baseline, Holm-corrected (N = 1,138 lines).
		Friedman omnibus test: $\chi^2 = 1978.68$, $p < 10^{-151}$.
		$\star$ = full fine-tuning baseline.
		*** $p < 0.001$, ** $p < 0.01$, * $p < 0.05$, ns = not significant ($p_{\text{Holm}} \ge 0.05$). VAN, GatedLexiconNet, and TrOCR base are external third-party systems included as reference points only; they are not TrOCR ablation variants and are not subject to statistical comparison.
	}
\end{table*}



\subsection{Ablation Analysis on I-CT 91}
We assess two common pipeline components---contrast normalization (CLAHE) and data augmentation---while keeping the model, split, and optimization fixed (full fine-tuning with \texttt{enc\_0\_dec\_0} under the same One-Cycle schedule). This allows us to attribute differences to the pipeline rather than to the training setup.
Our augmentation pipeline combines geometric, photometric, and degradation-style transforms. Rather than isolate each transform in a factorial design, we evaluate the full pipeline as a single intervention and compare it against a no-augmentation baseline.
On Cortonese, neither CLAHE nor the augmentation pipeline moves accuracy in a reliable direction once per-line variability is taken into account (Table~\ref{tab:ict91_ablation}). In the same setting, TrOCR is broadly comparable to the Kraken--CATMuS reference.

To study parameter-efficient adaptation, we progressively freeze larger prefixes of the encoder or decoder and compare configurations using line-level statistical testing (Table~\ref{tab:ict91_ablation}).
The outcome is markedly asymmetric. Encoder freezing is quickly detrimental once it extends beyond the first few layers. In contrast, the decoder is more tolerant: freezing a moderate fraction of decoder layers remains comparable to full fine-tuning, with clear degradation emerging only when most of the decoder is held fixed.
On Cortonese, the combined setting \texttt{enc\_3\_dec\_6} remains comparable to full fine-tuning while reducing the number of trainable parameters.

\subsection{Cross-Dataset Validation on READ-16}
We replicate the full ablation grid on READ-16 to test transfer beyond a single manuscript (Table~\ref{tab:read16_ablation}). READ-16 is used here as robustness evidence rather than as proof of universal generalization.

The CLAHE conclusion is stable: under the present optimization and model, contrast normalization does not emerge as a required ingredient. Augmentation, however, is more clearly beneficial on READ-16 than on Cortonese. 
The encoder--decoder asymmetry in freezing also persists, yet encoder freezing becomes less forgiving under this dataset shift, whereas decoder freezing remains comparatively safe up to a moderate depth.

The combined configuration \texttt{enc\_3\_dec\_6}  is non-significant on Cortonese but significantly worse than full fine-tuning on READ-16. 

\section{Discussion}

\subsection{Ablation and Baseline Results}
The freezing ablation yields a clear encoder--decoder asymmetry (Table~\ref{tab:ict91_ablation}). The encoder is the component that needs to learn the most: freezing beyond the first few layers is consistently harmful, which matches the intuition that the visual stack must adapt to manuscript-specific letterforms and degradation. The decoder, by contrast, is more transferable, and moderate decoder freezing can reduce training cost without a measurable loss on Cortonese.

The preprocessing ablation is equally instructive. On Cortonese, neither CLAHE nor the full augmentation bundle produces a reliable gain (Table~\ref{tab:ict91_ablation}). This suggests that, under One-Cycle training, TrOCR can learn robust features from minimally processed inputs, and that aggressive contrast enhancement is not a prerequisite for competitive accuracy. 

The READ-16 replication (Table~\ref{tab:read16_ablation}) sharpens the main cautions. The encoder remains the brittle part under domain shift, whereas the decoder continues to tolerate freezing up to a moderate depth. Augmentation, however, behaves differently and becomes clearly beneficial on READ-16, which is consistent with a dataset-dependent regularization effect. Finally, a combined freezing recipe that is acceptable on one corpus can become significantly worse on another, so freezing strategies should be re-validated per dataset rather than reused wholesale.

With respect to baselines, Kraken--CATMuS remains a strong reference point because it benefits from domain-relevant pre-training. The fact that TrOCR can reach a similar regime with careful fine-tuning and minimal preprocessing suggests that optimization choices can partially compensate for the lack of specialized pre-training, but it does not remove the motivation for domain-aware pre-training when such data are available.


\paragraph{Practical implications.} For TrOCR fine-tuning on low-resource historical line images, our results suggest four practical recommendations: 1) treat CLAHE as optional rather than default; 2) validate augmentation per dataset rather than assume a universal gain; 3) avoid aggressive encoder freezing; and 4) prefer full fine-tuning unless compute constraints justify moderate decoder freezing.

The ablations identify which fine-tuning choices change accuracy; the qualitative analysis is included to help interpret where these choices fail, especially in ambiguous or high-loss tokens. We therefore use Grad-CAM and cross-attention as diagnostic tools for understanding observed error modes.

\subsection{Diagnostic Interpretation of Ablation Results}

\begin{figure*}
	\centering
	\includegraphics[width=0.75\linewidth]{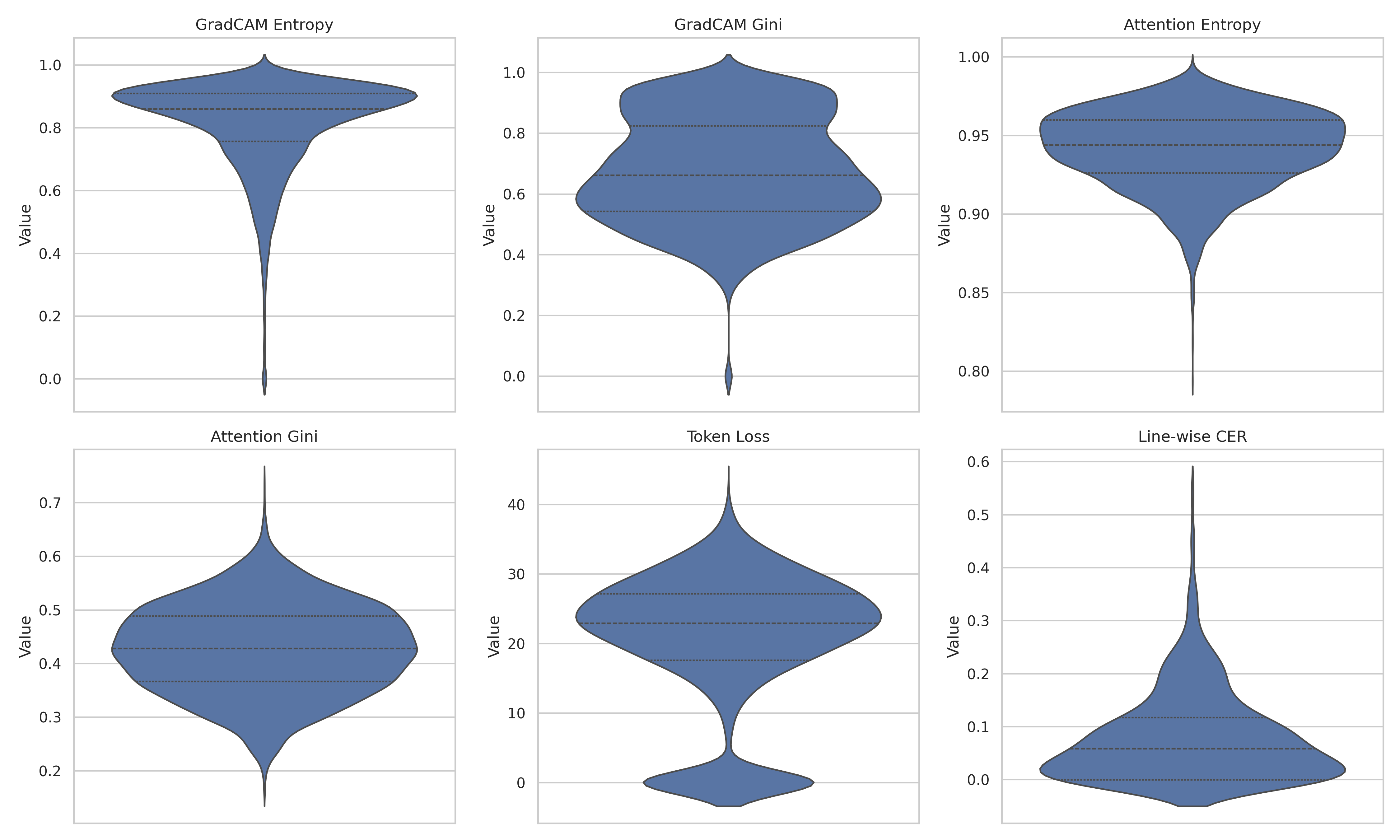}

	\caption{Distribution of metric values related to losses and Grad-CAM/attention maps. Violin plots and box plots are computed on the test set of the I-CT 91 dataset.}
	\label{fig:violins}
\end{figure*}
Normalized entropy is defined as:

\[
H' = \frac{-\sum_i p_i \log p_i}{\log(n)}
\]

where \(p_i\) denotes the normalized activation at patch \(i\), and \(n\) is the total number of patches. The normalized entropy satisfies \(H' \in [0,1]\) and measures how diffusely the attention map distributes its activation. A value of \(H' = 1\) corresponds to a uniform attention map with no spatial preference, whereas values of \(H'\) close to \(0\) indicate that the activation is concentrated on only a few patches.

In the context of HTR, a high-entropy map associated with an erroneous token suggests that the model lacks a confident spatial hypothesis for that character, making the metric diagnostically informative.
The Gini coefficient is defined as:

\[
G = \frac{2 \sum_{i=1}^{n} i\,p_i}{n \sum_{i=1}^{n} p_i} - \frac{n+1}{n}
\]

where the activations \(p_i\) are sorted in ascending order before computation, and \(n\) denotes the total number of patches. If the activations are normalized such that \(\sum_{i=1}^{n} p_i = 1\), the expression simplifies accordingly. The coefficient satisfies \(G \in [0,1]\), where values close to \(1\) indicate highly concentrated activation on a small number of patches, while values close to \(0\) correspond to a nearly uniform attention distribution.
Figure~\ref{fig:violins} summarizes Token Loss and two map statistics (normalized entropy and Gini) for both Grad-CAM and decoder cross-attention.
Because such aggregates can hide qualitative structure, we also inspected token-level overlays.
Table~\ref{tab:examples} shows four representative tokens spanning both low- and high-error lines, and highlights the recurring patterns and failure modes we observed.

Both maps are computed token-by-token from the final decoder block under teacher forcing.
For \emph{attention}, we average the cross-attention weights over heads, reshape them to the patch grid, and normalize.
For \emph{Grad-CAM}, we backpropagate the ground-truth logit for the same step, form a gradient-weighted sum over heads, apply ReLU, and normalize.
In both cases we drop the vision \texttt{[CLS]} patch.
Token Loss is computed on the same forward pass, and we summarize each map with normalized entropy and Gini.

\begin{table*}
	\centering
	\resizebox{\columnwidth}{!}{%
		\begin{tabular}{ccccp{0.55\linewidth}}
			\multirow{2}{*}{\textbf{\begin{tabular}[c]{@{}c@{}}Grad-CAM\\ Entr/Gini\end{tabular}}}  &
			\multirow{2}{*}{\textbf{\begin{tabular}[c]{@{}c@{}}Attention\\ Entr/Gini\end{tabular}}} &
			\multirow{2}{*}{\textbf{\begin{tabular}[c]{@{}c@{}}Token \\ Loss\end{tabular}}}         &
			\multirow{2}{*}{\textbf{\begin{tabular}[c]{@{}c@{}}Line \\ CER\end{tabular}}}           &
			\multicolumn{1}{c}{\multirow{2}{*}{\textbf{\begin{tabular}[c]{@{}c@{}}Grad-CAM (above) and\\\ Attention (below) maps\end{tabular}}}}                                                                                                                                                                                                                                             \\
			                                                                                        &   &  &  & \multicolumn{1}{c}{}                                                                                                                                                                                                                                                         \\ \hline
			0.97/0.31                                                                               &
			0.98/0.26                                                                               &
			24.5                                                                                    &
			39\%                                                                                    &
			\begin{minipage}{\linewidth} \centering \includegraphics[width=0.8\linewidth]{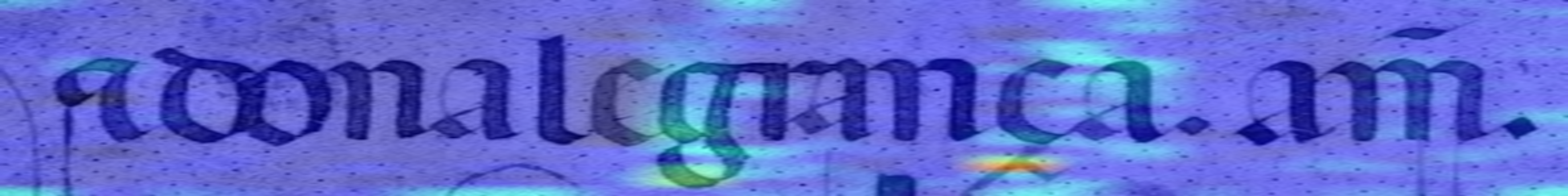} \vspace{2mm}\\ \includegraphics[width=0.8\linewidth]{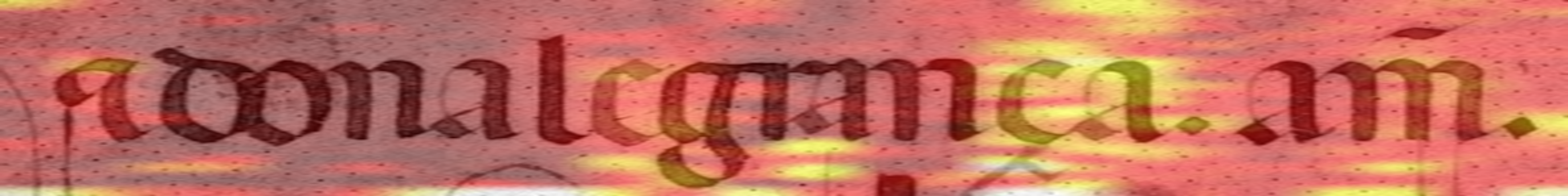} \vspace{2mm}\\ \small Token ``ran'' in ``et don' aleg{\ul ran}ça.''\\ TrOCR predicted ``et don alegrança a mer'' \end{minipage} \\ \hline
			0.35/0.99                                                                               &
			0.91/0.51                                                                               &
			19.3                                                                                    &
			0.0\%                                                                                   &
			\begin{minipage}{\linewidth} \centering \includegraphics[width=0.8\linewidth]{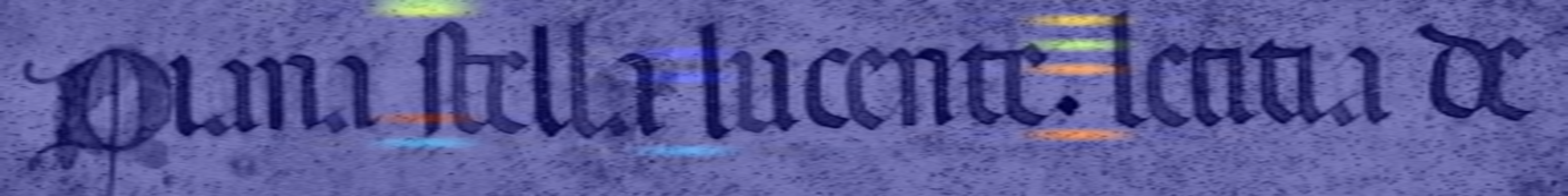} \vspace{2mm}\\ \includegraphics[width=0.8\linewidth]{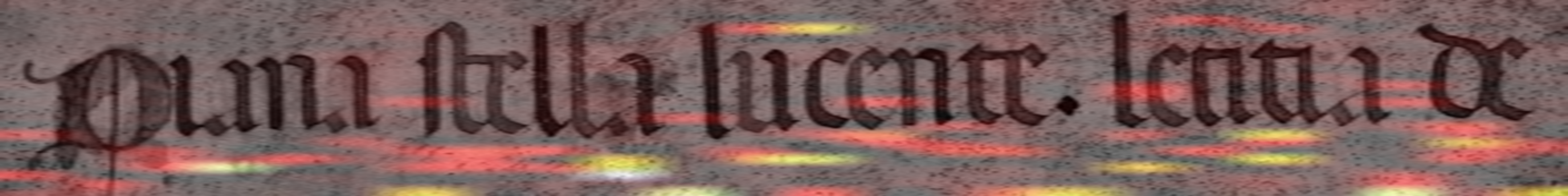} \vspace{2mm}\\ \small Token ``l'' in ``Diana stella {\ul l}ucente letitia de''\\ TrOCR predicted without errors. \end{minipage}     \\ \hline
			0.0/0.0                                                                                 &
			0.93/0.45                                                                               &
			35.3                                                                                    &
			0.0\%                                                                                   &
			\begin{minipage}{\linewidth} \centering \includegraphics[width=0.8\linewidth]{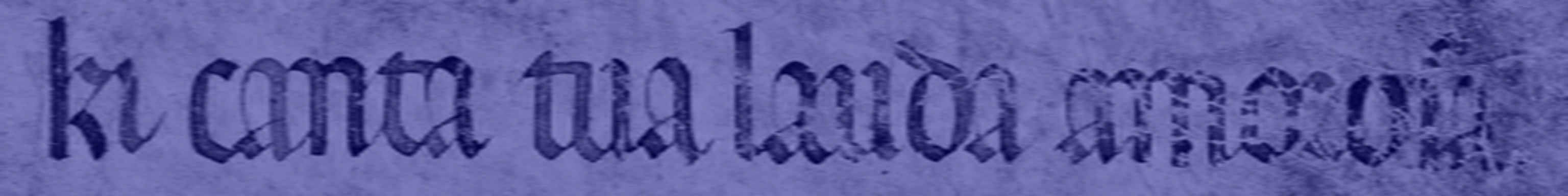} \vspace{2mm}\\ \includegraphics[width=0.8\linewidth]{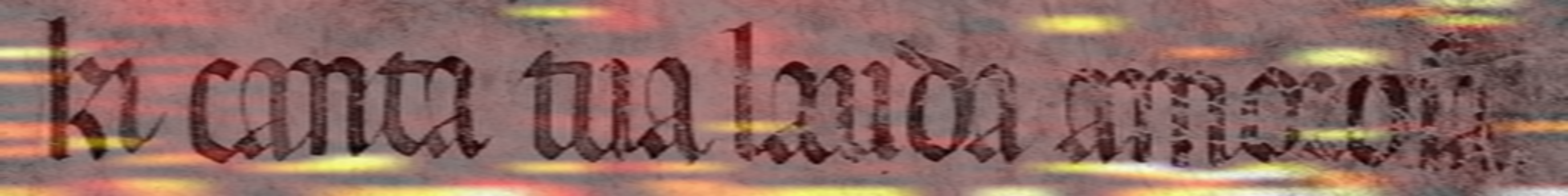} \vspace{2mm}\\ \small Token ``t'' in ``ki canta {\ul t}ua lauda amorosa''\\ TrOCR predicted without errors. \end{minipage}          \\ \hline
			0.77/0.71                                                                               &
			0.85/0.63                                                                               &
			12.2                                                                                    &
			0.0\%                                                                                   &
			\begin{minipage}{\linewidth} \centering \includegraphics[width=0.8\linewidth]{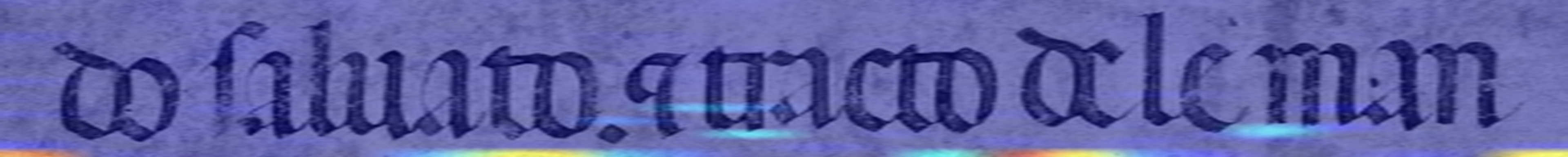} \vspace{2mm}\\ \includegraphics[width=0.8\linewidth]{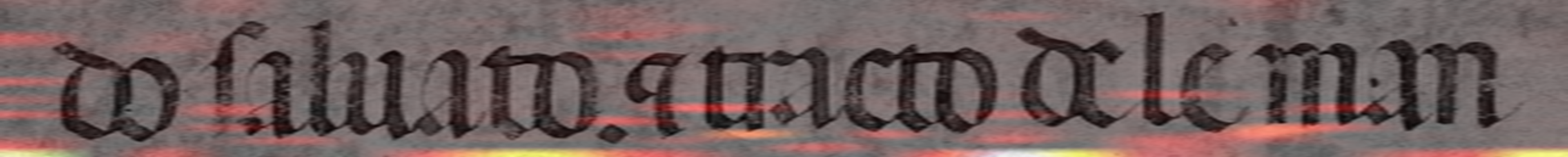} \vspace{2mm}\\ \small Token ``et'' in ``do salvato {\ul et} tracto de le man''\\ TrOCR predicted without errors. \end{minipage}   \\ \hline
		\end{tabular}%
	}
	\caption{Comparative analysis of Grad-CAM and attention summaries, Token Loss, and CER in four prediction scenarios. The first case pairs a high-error line with broadly distributed maps (high entropy, low Gini). In the second case, Grad-CAM is sharply localized while attention remains comparatively diffuse. In the third case, the Grad-CAM map collapses to (near-)zero. In the fourth case, attention is relatively more concentrated than Grad-CAM on a correctly recognized line.}
	\label{tab:examples}
\end{table*}

In the first two examples, Grad-CAM mostly follows token contours, emphasizing stroke boundaries more than filled interiors.
This ``edge-seeking'' behavior is consistent with the model relying on high-frequency ink--background transitions, and with Grad-CAM highlighting precisely the regions to which the token-level loss is most sensitive.
Cross-attention is often less crisp in the overlays: even after averaging across heads, it can remain spread over nearby patches, including in correctly decoded cases, which suggests that the decoder may maintain multiple plausible alignments rather than committing to a single peak.

The metrics in Table~\ref{tab:examples} align with these visual regimes.
When the line is difficult and the token loss is high, both maps tend to have high entropy and low Gini, indicating broad, weakly concentrated support.
Conversely, a line can be transcribed correctly while still containing tokens with elevated loss; in such cases, Grad-CAM may remain sharply localized even when attention looks comparatively diffuse.
This is a useful reminder that ``correct'' is not the same as ``confident'', and that the two map types capture different aspects of the decoding process.

The third row highlights a practical caveat: Grad-CAM can collapse to an all-zero map, making entropy and Gini uninformative by construction.
We treat this as a failure mode of the signal (e.g., negative contributions removed by ReLU or vanishing gradients), not as evidence of extreme certainty.
In these cases, attention remains well-defined, so attention-derived diagnostics can still provide a fallback.

The fourth row shows the complementary situation, where attention is relatively more concentrated than Grad-CAM on a correctly recognized line.

Taken together, the examples reinforce that attention and Grad-CAM answer different questions.
For downstream diagnostics, this motivates a simple hybrid rule: use Grad-CAM-based measures when the map is non-degenerate, and fall back to attention-based measures (or explicitly flag the token) when Grad-CAM energy is near zero.

\section{Conclusion and Future Work}

This paper presents a controlled ablation study of practical TrOCR fine-tuning choices for low-resource medieval HTR. Across Cortonese and READ-16, the most stable findings are that encoder freezing is more fragile than decoder freezing, CLAHE is not required under the present recipe, and augmentation and combined freezing strategies remain dataset-dependent.


On Cortonese, CLAHE and our augmentation bundle are optional under this recipe: neither produces a statistically reliable gain (Table~\ref{tab:ict91_ablation}). Freezing behaves differently. The encoder is the brittle part: freezing past \texttt{enc\_3} quickly becomes harmful, whereas the decoder remains tolerant up to \texttt{dec\_6}. Under these constraints, \texttt{enc\_3\_dec\_6} remains non-significant with respect to full fine-tuning.

READ-16 preserves the broad asymmetry (decoder more transferable than encoder) and again does not elevate CLAHE to a required component (Table~\ref{tab:read16_ablation}). What changes is that augmentation becomes beneficial, and the combined \texttt{enc\_3\_dec\_6} recipe turns significantly worse than full fine-tuning. In other words, encoder freezing thresholds and, especially, combined freezes do not automatically carry over across corpora.


Beyond accuracy, Grad-CAM and decoder cross-attention are useful as diagnostic tools for interpreting ablation-linked errors, but they are best treated as complementary evidence alongside sequence-level signals such as token loss.

These conclusions are strongest for TrOCR-like encoder--decoder HTR models on low-resource historical line images. They should not be overgeneralized to all HTR architectures or datasets, and READ-16 should be interpreted as robustness evidence rather than proof of universal transfer. Even so, the study can help future researchers avoid redundant ablation work when adapting TrOCR in similar settings. These conclusions should be validated on additional scripts, periods, datasets, and HTR architectures.

Future work is largely methodological. The most direct next step is domain-aware pre-training on paleographic corpora (e.g., CATMuS-like collections), followed by the same cross-dataset evaluation protocol, to separate ``better initialization'' from ``better fine-tuning.'' In parallel, uncertainty estimation should be made robust to edge cases by masking structural tokens (\texttt{<s>}, \texttt{</s>}, \texttt{[PAD]}), detecting degenerate Grad-CAM maps, and combining map-based summaries with calibrated sequence-level signals. These signals are useful only if they hold up downstream, so we consider active learning and semi-supervised experiments---measured as annotation savings at a fixed target quality---to be the next required validation. Finally, freezing is not the only parameter-efficient option; comparing it against alternatives such as low-rank adapters, and replicating across scripts, languages, and material conditions, would make the results more generalizable.

\section*{Funding Statement}
\begin{wrapfigure}{l}{0.35\linewidth}
    \vspace{-\intextsep} 
    \vspace{5pt} 
    \includegraphics[width=0.95\linewidth]{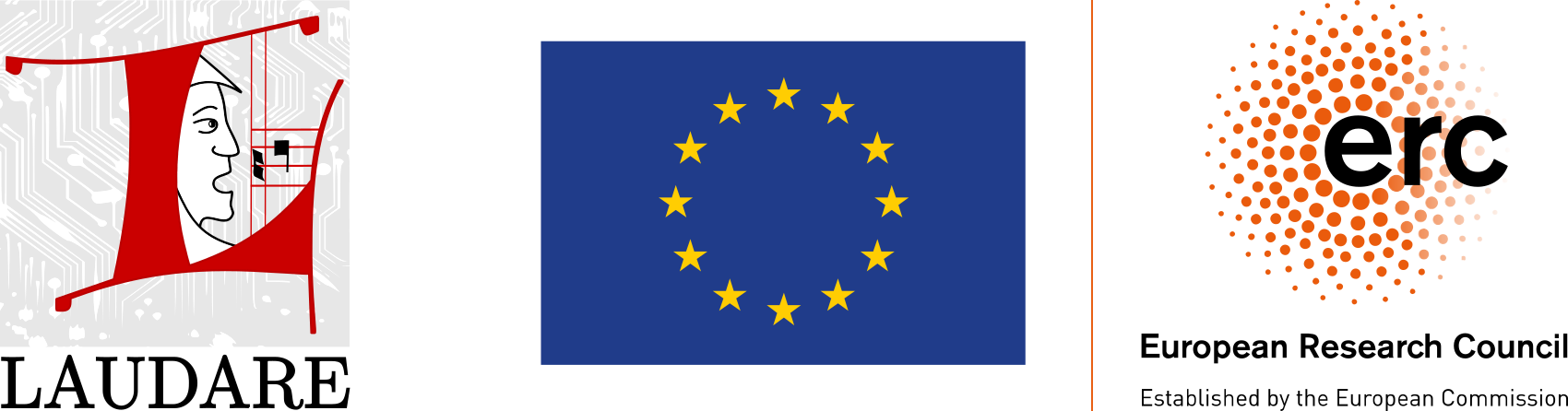}
    \vspace{-\intextsep} 
    \vspace{5pt} 
\end{wrapfigure} 
This work has been funded by the European Union (Horizon Programme for Research and Innovation 2021-2027, ERC Advanced Grant ``The Italian Lauda: Disseminating Poetry and Concepts Through Melody (12th-16th century)'', acronym LAUDARE, project no. 101054750). The views and opinions expressed are, however, only those of the author and do not necessarily reflect those of the European Union or the European Research Council. Neither the European Union nor the awarding authority can be held responsible for such matters.

\bibliographystyle{splncs04nat}
\bibliography{references,related_works}

@inproceedings{clerice2024catmus,
  title = {{{CATMuS}} Medieval: A Multilingual Large-Scale Cross-Century Dataset in~Latin Script for~Handwritten Text Recognition and~Beyond},
  shorttitle = {{{CATMuS}} Medieval},
  booktitle = {Document {{Analysis}} and {{Recognition}} - {{ICDAR}} 2024},
  author = {Cl{\'e}rice, Thibault and Pinche, Ariane and {Vlachou-Efstathiou}, Malamatenia and Chagu{\'e}, Alix and Camps, Jean-Baptiste and Levenson, Matthias Gille and {Brisville-Fertin}, Olivier and Boschetti, Federico and Fischer, Franz and Gervers, Michael and Boutreux, Agn{\`e}s and Manton, Avery and Gabay, Simon and O'Connor, Patricia and Haverals, Wouter and Kestemont, Mike and Vandyck, Caroline and Kiessling, Benjamin},
  editor = {Barney Smith, Elisa H. and Liwicki, Marcus and Peng, Liangrui},
  year = 2024,
  pages = {174--194},
  publisher = {Springer Nature Switzerland},
  address = {Cham},
  doi = {10.1007/978-3-031-70543-4_11},
}

@inproceedings{sharma2025experimenting,
  title = {Experimenting Active and Sequential Learning in a Medieval Music Manuscript},
  booktitle = {{{IEEE International Workshop}} on {{Machine Learning}} for {{Signal Processing}}},
  author = {Sharma, Sachin and Simonetta, Federico and Flammini, Michele},
  year = 2025,
  month = jul,
  eprint = {2507.15633},
  primaryclass = {cs},
  publisher = {arXiv},
  address = {Istanbul},
  doi = {10.48550/arXiv.2507.15633},
}

@inproceedings{simonetta2024optical,
  title = {Optical Music Recognition in Manuscripts from the Ricordi Archive},
  booktitle = {{{AudioMostly}} '24},
  author = {Simonetta, Federico and Mondal, Rishav and Ludovico, Luca Andrea and Ntalampiras, Stavros},
  year = 2024,
  month = aug,
  eprint = {2408.10260},
  primaryclass = {cs},
  publisher = {ACM},
  address = {Milan, Italy},
  doi = {10.1145/3678299.3678324},
}

@misc{kumari2024gatedlexiconnet,
  title = {{{GatedLexiconNet}}: A Comprehensive End-to-End Handwritten Paragraph Text Recognition System},
  shorttitle = {{{GatedLexiconNet}}},
  author = {Kumari, Lalita and Singh, Sukhdeep and Rathore, Vaibhav Varish Singh and Sharma, Anuj},
  year = 2024,
  month = apr,
  number = {arXiv:2404.14062},
  eprint = {2404.14062},
  primaryclass = {cs},
  publisher = {arXiv},
  doi = {10.48550/arXiv.2404.14062},
  urldate = {2025-11-12},
  archiveprefix = {arXiv},
  langid = {english},
}

@inproceedings{huttner2025lowrank,
  title = {Low-Rank Adaptation vs. {{Fine-tuning}} for Handwritten Text Recognition},
  booktitle = {2025 {{IEEE}}/{{CVF Winter Conference}} on {{Applications}} of {{Computer Vision Workshops}} ({{WACVW}})},
  author = {H{\"u}ttner, Lukas and Mayr, Martin and Gorges, Thomas and Wu, Fei and Seuret, Mathias and Maier, Andreas and Christlein, Vincent},
  year = 2025,
  month = feb,
  pages = {1233--1242},
  issn = {2690-621X},
  doi = {10.1109/WACVW65960.2025.00146},
  urldate = {2025-11-12},
}

@inproceedings{kiessling2026version,
  title = {Version 5 of~the~Kraken {{ATR}} Engine for~the~Humanities},
  booktitle = {Document {{Analysis}} and {{Recognition}} -- {{ICDAR}} 2025},
  author = {Kiessling, Benjamin},
  editor = {Yin, Xu-Cheng and Karatzas, Dimosthenis and Lopresti, Daniel},
  year = 2026,
  pages = {443--458},
  publisher = {Springer Nature Switzerland},
  address = {Cham},
  doi = {10.1007/978-3-032-04624-6_26},
}

@misc{pinche_2024_12743230,
  author       = {Pinche, Ariane and
                  Clérice, Thibault and
                  Chagué, Alix and
                  Camps, Jean-Baptiste and
                  Vlachou-Efstathiou, Malamatenia and
                  Gille Levenson, Matthias and
                  Brisville-Fertin, Olivier and
                  Boschetti, Federico and
                  Fischer, Franz and
                  Gervers, Michael and
                  Boutreux, Agnès and
                  Manton, Avery and
                  Gabay, Simon},
  title        = {CATMuS Medieval},
  month        = jul,
  year         = 2024,
  publisher    = {Zenodo},
  version      = {1.5.0},
  doi          = {10.5281/zenodo.12743230},
}

@inproceedings{selvarajuGradcamVisualExplanations2017,
  title = {Grad-Cam: {{Visual Explanations}} from {{Deep Networks Via Gradient-based Localization}}},
  booktitle = {Proc. {{IEEE Int}}. {{Conf}}. {{Computer Vision}} ({{ICCV}})},
  author = {Selvaraju, R. R. and Cogswell, M. and Das, A. and Vedantam, R. and Parikh, D. and Batra, D.},
  year = 2017,
  month = oct,
  pages = {618--626},
  issn = {2380-7504},
  doi = {10.1109/ICCV.2017.74}
}

@article{coquenet2023endtoend,
  title = {End-to-End Handwritten Paragraph Text Recognition Using a Vertical Attention Network},
  author = {Coquenet, Denis and Chatelain, Cl{\'e}ment and Paquet, Thierry},
  year = 2023,
  month = jan,
  journal = {IEEE Transactions on Pattern Analysis and Machine Intelligence},
  volume = {45},
  number = {1},
  pages = {508--524},
  issn = {1939-3539},
  doi = {10.1109/TPAMI.2022.3144899},
  urldate = {2025-11-12},
}

@inproceedings{li2021trocr,
title = {{{TrOCR}}: Transformer-Based Optical Character Recognition with Pre-Trained Models},
  shorttitle = {{{TrOCR}}},
  booktitle = {Proceedings of the {{AAAI Conference}} on {{Artificial Intelligence}}},
  author = {Li, Minghao and Lv, Tengchao and Chen, Jingye and Cui, Lei and Lu, Yijuan and Florencio, Dinei and Zhang, Cha and Li, Zhoujun and Wei, Furu},
  year = 2023,
  month = jun,
  volume = {37},
  pages = {13094--13102},
  doi = {10.1609/aaai.v37i11.26538},
  urldate = {2025-11-11},
}

@misc{strobel2022transformerbasedhtrhistoricaldocuments,
  title={Transformer-based HTR for Historical Documents},
  author={Phillip Benjamin Ströbel and Simon Clematide and Martin Volk and Tobias Hodel},
  year={2022},
  eprint={2203.11008},
  archivePrefix={arXiv},
  primaryClass={cs.CV},
  url={https://arxiv.org/abs/2203.11008}
}

@inproceedings{chefer2021transformer,
  title={Transformer Interpretability Beyond Attention Visualization},
  author={Chefer, Hila and Gur, Shir and Wolf, Lior},
  booktitle={Proceedings of the IEEE/CVF Conference on Computer Vision and Pattern Recognition (CVPR)},
  pages={782–791},
  year={2021},
  doi={10.1109/CVPR46437.2021.00083}
}

@inbook{10.5555/180895.180940,
  author = {Zuiderveld, Karel},
  title = {Contrast limited adaptive histogram equalization},
  year = {1994},
  isbn = {0123361559},
  publisher = {Academic Press Professional, Inc.},
  address = {USA},
  booktitle = {Graphics Gems IV},
  pages = {474--485},
  numpages = {12}
}

@article{Pizer1987AdaptiveHE,
  title={Adaptive histogram equalization and its variations},
  author={Stephen M. Pizer and Elton Philip Amburn and John D. Austin and Robert Cromartie and Ari Geselowitz and Trey Greer and Bart M. ter Haar Romeny and John B. Zimmerman},
  journal={Computer Vision, Graphics, and Image Processing},
  year={1987},
  volume={39},
  pages={355--368},
  url={https://api.semanticscholar.org/CorpusID:62771950}
}

@inproceedings{Xu_2020,
  author    = {Xu, Yiheng and Li, Minghao and Cui, Lei and Huang, Shaohan and Wei, Furu and Zhou, Ming},
  title     = {LayoutLM: Pre-training of Text and Layout for Document Image Understanding},
  booktitle = {Proceedings of the 26th ACM SIGKDD International Conference on Knowledge Discovery \& Data Mining},
  series    = {KDD '20},
  pages     = {1192--1200},
  year      = {2020},
  month     = aug,
  publisher = {ACM},
  doi       = {10.1145/3394486.3403172},
}

@article{Fischer2023,
  author  = {Fischer, N. and Hartelt, A. and Puppe, F.},
  title   = {Line-Level Layout Recognition of Historical Documents with Background Knowledge},
  journal = {Algorithms},
  year    = {2023},
  volume  = {16},
  number  = {3},
  pages   = {136},
  doi     = {10.3390/a16030136},
}

@article{Kim2021Donut,
  author  = {Kim, Geewook and Hong, Teakgyu and Yim, Moonbin and Park, Jinyoung and Yim, Jinyeong and Hwang, Wonseok and Yun, Sangdoo and Han, Dongyoon and Park, Seunghyun},
  title   = {Donut: Document Understanding Transformer without OCR},
  journal = {arXiv preprint arXiv:2111.15664},
  year    = {2021},
  volume  = {7},
  number  = {15},
  pages   = {2},
  url     = {https://arxiv.org/abs/2111.15664}
}

@misc{Toselli2018READ,
  author       = {Toselli, Alejandro H. and Romero, Victor and Villegas, Manuel and Vidal, Enrique and Sánchez, Javier A.},
  title        = {HTR Dataset ICFHR 2016 [Data set]},
  year         = {2018},
  howpublished = {Zenodo},
  doi          = {10.5281/zenodo.1164045},
}

@inproceedings{Fischer2011SaintGall,
  author       = {Fischer, Andreas and Frinken, Volkmar and Fornés, Alicia and Bunke, Horst},
  title        = {Transcription alignment of Latin manuscripts using hidden Markov models},
  booktitle    = {Proceedings of the 2011 Workshop on Historical Document Imaging and Processing},
  series       = {HIP '11},
  pages        = {29--36},
  year         = {2011},
  location     = {Beijing, China, USA},
  publisher    = {Association for Computing Machinery},
  doi          = {10.1145/2037342.2037348},
}

@inproceedings{Toselli2015Bentham,
  author       = {Toselli, Alejandro H. and Vidal, Enrique},
  title        = {Handwritten Text Recognition Results on the Bentham Collection with Improved Classical N-Gram-HMM methods},
  booktitle    = {Proceedings of the 3rd International Workshop on Historical Document Imaging and Processing},
  series       = {HIP '15},
  pages        = {15--22},
  year         = {2015},
  location     = {Gammarth, Tunisia},
  publisher    = {Association for Computing Machinery},
  doi          = {10.1145/2809544.2809551},
}

@incollection{semanticscholar110,
  title = {{{OCR-free}} Document Understanding Transformer},
  booktitle = {European {{Conference}} on {{Computer Vision}}},
  author = {Kim, Geewook and Hong, Teakgyu and Yim, Moonbin and Nam, JeongYeon and Park, Jinyoung and Yim, Jinyeong and Hwang, Wonseok and Yun, Sangdoo and Han, Dongyoon and Park, Seunghyun},
  editor = {Avidan, Shai and Brostow, Gabriel and Ciss{\'e}, Moustapha and Farinella, Giovanni Maria and Hassner, Tal},
  year = 2021,
  langid = {english}
}

@incollection{semanticscholar111,
  title = {On Text Localization in End-to-End {{OCR-free}} Document Understanding Transformer without Text Localization Supervision},
  booktitle = {{{ICDAR Workshops}}},
  author = {Kim, Geewook and Yokoo, Shuhei and Seo, Sukmin and Osanai, Atsuki and Okamoto, Yamato and Baek, Youngmin},
  editor = {Coustaty, Mickael and Forn{\'e}s, Alicia},
  year = 2023,
  langid = {english}
}

@incollection{semanticscholar14,
  title = {A Historical Handwritten Dataset for Ethiopic {{OCR}} with Baseline Models and Human-Level Performance},
  booktitle = {{{IEEE International Conference}} on {{Document Analysis}} and {{Recognition}}},
  author = {Belay, Birhanu Hailu and Guyon, Isabelle and Mengiste, Tadele and Tilahun, Bezawork and Liwicki, Marcus and Tegegne, Tesfa and Egele, Romain},
  editor = {Barney Smith, Elisa H. and Liwicki, Marcus and Peng, Liangrui},
  year = 2024,
  langid = {english}
}

@incollection{semanticscholar20,
  title = {The Adaptability of a Transformer-Based {{OCR}} Model for Historical Documents},
  booktitle = {{{ICDAR Workshops}}},
  author = {Str{\"o}bel, Phillip Benjamin and Hodel, Tobias and Boente, Walter and Volk, Martin},
  editor = {Coustaty, Mickael and Forn{\'e}s, Alicia},
  year = 2023,
  langid = {english}
}

@misc{semanticscholar31,
  title = {{{TRIDIS}}: A Comprehensive Medieval and Early Modern Corpus for {{HTR}} and {{NER}}},
  author = {Aguilar, Sergio Torres},
  year = 2025,
  doi = {10.48550/arXiv.2503.22714},
  langid = {english}
}

@article{semanticscholar32,
  title = {{{iForal}}: Automated Handwritten Text Transcription for Historical Medieval Manuscripts},
  shorttitle = {{{iForal}}},
  author = {Matos, Alexandre and Almeida, Pedro and Correia, Paulo and Pacheco, Osvaldo},
  year = 2025,
  month = jan,
  journal = {J. Imaging},
  volume = {11},
  number = {2},
  pages = {36},
  issn = {2313-433X},
  doi = {10.3390/jimaging11020036},
  urldate = {2025-11-11},
  langid = {english}
}

@article{semanticscholar41,
  title = {From Manuscript to Metadata: Experiments on Handwritten Text Recognition, Tagging and Importation for the Memoriali Series (1265-1452)},
  shorttitle = {From Manuscript to Metadata},
  author = {Loss, Edward and Guernaccini, Fabiana and Carassai, Manuel},
  year = 2025,
  month = may,
  journal = {Jlis.it},
  volume = {16},
  number = {2},
  pages = {59--85},
  issn = {2038-1026},
  doi = {10.36253/jlis.it-641},
  urldate = {2025-11-11},
  copyright = {https://creativecommons.org/licenses/by/4.0},
  langid = {english}
}

@misc{semanticscholar46,
  title = {Handwritten Text Recognition of Historical Manuscripts Using Transformer-Based Models},
  author = {Meoded, Erez},
  year = 2025,
  month = aug,
  doi = {10.48550/arXiv.2508.11499},
  langid = {english}
}

\end{document}